\documentclass{article}
\usepackage{spconf,amsmath,graphicx,hyperref,xcolor,amssymb,booktabs,multirow,adjustbox,dblfloatfix,float,placeins}
\usepackage[caption=false,font=footnotesize]{subfig}
\usepackage{bbm}
\usepackage{xcolor}

\title{Local2Mesh: Spatially Localized Contour-to-Mesh for Left Ventricular Reconstruction from Sparse 2D Cardiac MRI}
\name{
Haoyu Wu$^{1}$,
Ling Lin$^{2}$,
Pascal Lefèvre$^{1}$,
Ruizhe Li$^{3}$,
Xiaowu Sun$^{1}$
}

\address{\fontsize{11.5}{13}\selectfont
$^{1}$ Academy of Artificial Intelligence and Advanced Technology,
Xi'an Jiaotong-Liverpool University, China \\
\fontsize{11.5}{13}\selectfont
$^{2}$ Department of Health Technology and Informatics, The Hong Kong Polytechnic University, China \\
\fontsize{11.5}{13}\selectfont
$^{3}$ Nottingham Biomedical Research Centre, School of Medicine, University of Nottingham, UK
}
\begin{document}
%
\maketitle
\begin{abstract}
Three-dimensional (3D) left ventricular (LV) reconstruction from sparse cardiac magnetic resonance (CMR) imaging remains challenging due to inter-slice misalignment and insufficient local spatial information between slices. Global aggregation of contour features may obscure local contour-to-surface relationships. We propose Local2Mesh, a spatially localized contour-to-mesh framework that deforms a template mesh to reconstruct 3D LV geometry from sparse 2D contours without 3D mesh annotations. The framework introduces geometry-aware alignment to correct inter-slice misalignment and a plane-aware Local Router that routes contour features to template vertices using vertex-to-plane distances. Local and global contour features then jointly guide graph-based template deformation for 3D LV reconstruction. Experiments on two public datasets, M\&Ms-2 and ACDC, demonstrate superior geometric reconstruction and functional estimation over existing methods. Zero-shot transfer from M\&Ms-2 to ACDC demonstrates strong cross-dataset generalization. Reconstructed meshes also improve disease classification over sparse contours, supporting their utility for downstream cardiac analysis. These results demonstrate that combining geometry-aware alignment with local contour-to-vertex modeling improves LV reconstruction from sparse 2D contours and supports downstream cardiac analysis.
The code is available at \url{https://github.com/hwu918945-alt/loca2mesh}.
\end{abstract}



\vspace{-4.5mm}
\section{Introduction}
\label{sec:intro}
\vspace{-2mm}

Cardiac magnetic resonance (CMR) imaging is the reference standard for quantitative assessment of cardiac structure and function. Cine CMR typically covers the left ventricle (LV) with a stack of two-dimensional (2D) short-axis (SAX) slices, complemented by long-axis (LAX) views. These spatially sparse slices incompletely represent the three-dimensional (3D) anatomy. Slices acquired during separate breath-holds can also be spatially misaligned, further compromising the geometric consistency among the observed slices\cite{beetz2023pccn,meng2022mulvimotion}. 
Reconstructing complete LV geometry therefore requires both inferring unobserved anatomy and resolving inter-slice misalignment. In this work, we aim to reconstruct patient-specific 3D LV geometry from sparse CMR imaging at a single cardiac phase.

Several approaches have been proposed to reconstruct 3D cardiac geometry from sparse 2D CMR slices. Statistical shape models impose anatomical priors to infer unobserved geometry, but constrain reconstruction to a predefined shape space~\cite{yang2025adalvmf}. Learning-based methods infer dense geometry from sparse contour point clouds, but often rely on synthetic or densely sampled 3D shapes for supervision~\cite{beetz2023pccn}. More recent methods reconstruct 3D cardiac geometry from sparse 2D observations without dense 3D supervision. GHDHeart fits patient-specific geometry through compact global graph-harmonic deformation~\cite{luo2026ghdheart}. Slice2Mesh deforms a template mesh using sparse 2D slice supervision~\cite{xiao2025slice2mesh}, while TetHeart incorporates slice-adaptive features to reconstruct dynamic cardiac geometry~\cite{chen2025tetheart}. Despite these advances, existing methods do not explicitly account for the local spatial relevance of observed contours to different surface locations. Each contour directly constrains the anatomical boundary within its acquisition plane. Under a local smoothness assumption, this boundary can also guide surface reconstruction in nearby unobserved regions. This motivates explicitly modelling the spatial relationship between each contour and the location being reconstructed when inferring continuous 3D surfaces from sparse 2D slices.


We develop Local2Mesh, a patient-specific 3D LV reconstruction framework that addresses inter-slice misalignment and explicitly models the spatial relationship between each contour and the location being reconstructed. It reconstructs geometry at end-diastole (ED) or end-systole (ES) using only 2D contour supervision. Our contributions are threefold:

\par
\begingroup
\setlength{\parskip}{0pt}

\noindent\hspace*{1em}\textbullet\enspace
To address inter-slice misalignment, we propose a geometry-informed approach applicable across acquisition settings.
\par
\noindent\hspace*{1em}\textbullet\enspace
To preserve the spatial locality of contour features, we introduce plane-aware contour-to-vertex routing that integrates global anatomical context to guide template deformation.
\par
\noindent\hspace*{1em}\textbullet\enspace
We demonstrate robust 3D LV reconstruction using only 2D contour supervision across cine CMR datasets.
\par
\endgroup

\vspace{-4mm}
\section{METHOD}
\label{sec:METHOD}

\begin{figure*}[t]
    \centering
    \includegraphics[
        width=0.94\textwidth,
        trim=0 5mm 0 0,
        clip
    ]{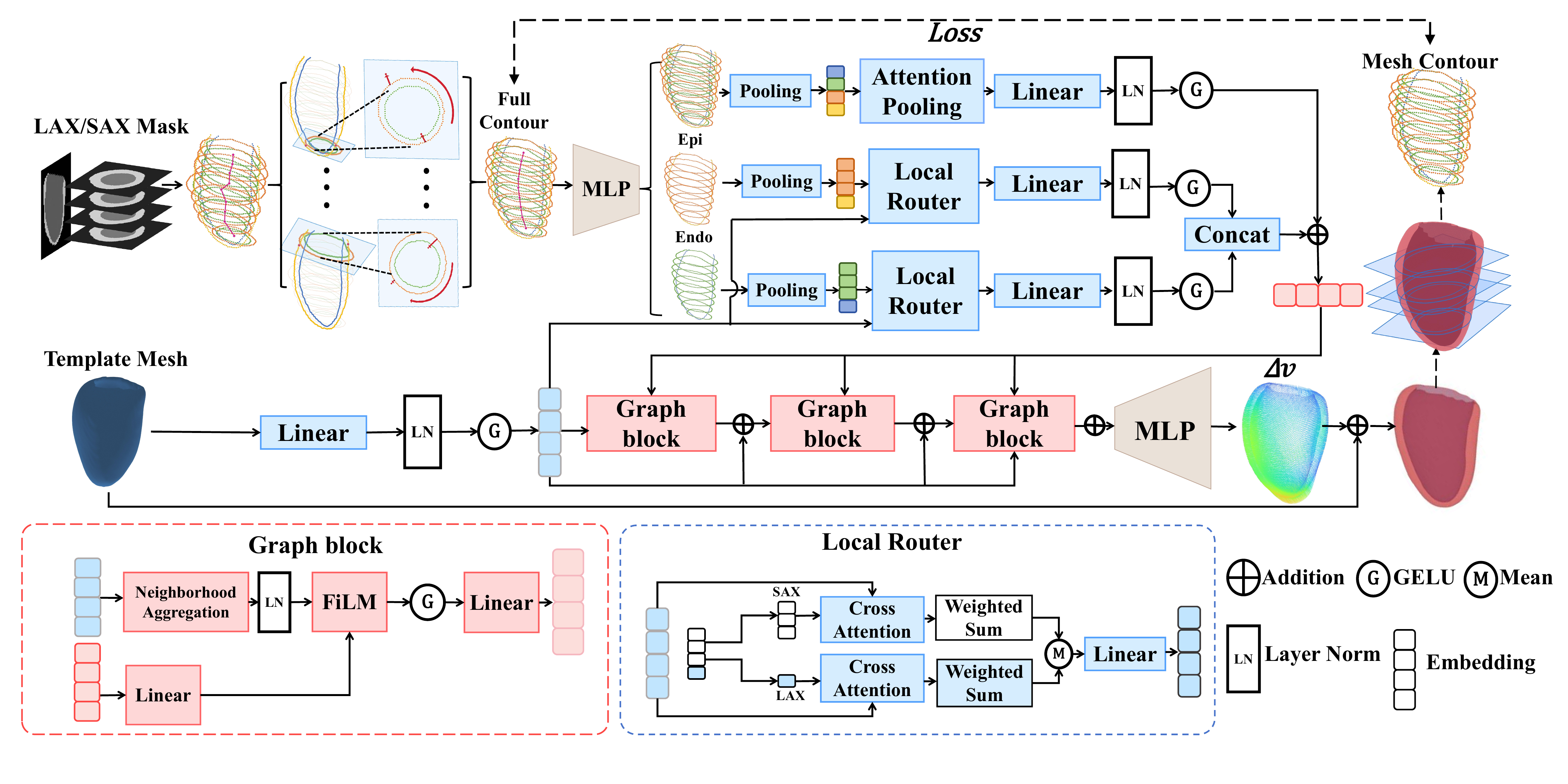}
    \vspace{-7mm}
    \caption{Overview of Local2Mesh. Aligned 2D contours are encoded into global and surface-specific features, which are routed to template vertices and integrated to condition graph-based deformation for patient-specific LV reconstruction.}
    \label{fig:framework}
\end{figure*}
\vspace{-2mm}
As shown in Fig.~\ref{fig:framework}, Local2Mesh first aligns 2D contours in 3D space, then integrates global and vertex-wise local features to guide graph-based template deformation, and finally reconstructs patient-specific LV geometry. We next describe contour alignment, the reconstruction network, and 2D contour-based training loss.
\vspace{-4mm}

\subsection{Geometric Contour Alignment}
\vspace{-1mm}
For each SAX and LAX slice, endocardial and epicardial contours are extracted from the segmentation masks and positioned in 3D space according to the corresponding imaging plane. We denote the contour set by $\mathcal C=\bigcup_{r,m}\mathcal C^{r,m}$, where $r\in\{\mathrm{epi},\mathrm{endo}\}$ and $m\in\{\mathrm{SAX},\mathrm{LAX}\}$ indicate the anatomical boundary and acquisition view, respectively. Let $\mathcal C_i^{\mathrm{epi}}\in\mathcal C^{\mathrm{epi,SAX}}$ denote the epicardial contour of SAX slice $i$, with center $\mathbf c_i$. To correct inter-slice motion, we estimate an in-plane translation $\mathbf t_i\in\mathbb R^2$, yielding the translated center $\mathbf c'_i=\mathbf c_i+\mathbf t_i$. A quadratic curve is fitted to the translated centers along the normalized stack position, and $\mathbf c'_i$ is constrained toward its fitted position $\hat{\mathbf c}_i$ by
\vspace{-2mm}
\begin{equation}
\mathcal L_{\mathrm{center}}
=
\frac{2}{N}\sum_{i=1}^{N}
\left[
\sqrt{1+
\left(\|\mathbf c'_i-\hat{\mathbf c}_i\|_2/s_c\right)^2}
-1
\right],
\label{eq:alignment_center}
\end{equation}
where $N$ is the number of valid SAX slices and $s_c>0$ is the robust scale parameter. The translations are additionally regularized toward a quadratic trajectory along the stack, with a magnitude penalty to prevent excessive displacement.

When LAX contours are available, their intersections with SAX plane $\Pi_i$ define two anchors $\mathbf h_{i,1}$ and $\mathbf h_{i,2}$. The alignment is then extended to an in-plane rigid transformation $T_i$, parameterized by rotation $\theta_i$ and translation $\mathbf t_i$. The transformed contour $\mathcal C_i^{\prime,\mathrm{epi}}=T_i(\mathcal C_i^{\mathrm{epi}})$ is constrained to pass through these anchors by
\vspace{-2mm}
\begin{equation}
\mathcal L_{\mathrm{hit}}
=
\frac{1}{|\mathcal H|}
\sum_{i\in\mathcal H}\sum_{a=1}^{2}
\left[
\sqrt{1+
\left(
\operatorname{dist}(\mathbf h_{i,a},\mathcal C_i^{\prime,\mathrm{epi}})/s_h
\right)^2}
-1
\right],
\label{eq:alignment_hit}
\end{equation}
where $\mathcal H$ contains SAX planes with two valid intersections, $\operatorname{dist}(\cdot,\cdot)$ is the shortest point-to-contour distance, and $s_h>0$ is the robust scale parameter. Since the two anchors do not fully determine the in-plane pose, the center constraint is retained and rotations are regularized across the SAX stack. Alignment is performed independently for each cardiac phase without ground-truth transformations.

\vspace{-3mm}
\subsection{Network Structure}
\vspace{-1mm}
As illustrated in Fig.~\ref{fig:framework}, the reconstruction network takes the aligned contour set $\mathcal C$ and a fixed LV template mesh $\mathcal G=(\mathcal V,\mathcal E)$ as input. The template is adopted from the statistical shape model~\cite{bai2015atlas}, where $\mathcal V=\{\mathbf v_i\}_{i=1}^{N_v}$ contains $N_v$ vertices and $\mathcal E$ denotes the mesh connectivity. Given $\mathcal C$ and $\mathcal G$, the reconstruction network
$F_\Theta$ predicts vertex displacements:

\vspace{-3mm}
{
\setlength{\abovedisplayskip}{0pt}
\setlength{\belowdisplayskip}{0pt}
\setlength{\abovedisplayshortskip}{0pt}
\setlength{\belowdisplayshortskip}{0pt}
\begin{equation}
\{\Delta\mathbf v_i\}_{i=1}^{N_v}
=
F_\Theta(\mathcal C,\mathcal G),
\quad
\hat{\mathbf v}_i
=
\mathbf v_i+\Delta\mathbf v_i,
\label{eq:mesh_reconstruction}
\end{equation}
}
\vspace{-1mm}
where $\hat{\mathbf v}_i$ denotes the reconstructed position of vertex $i$. The network encodes contour features, routes local contour information to individual template vertices, and deforms the template through graph-based conditioning, as detailed below.


\vspace{-3mm}

\subsubsection{Contour Feature Encoding}

The full contour $\mathcal C$ is processed by a point-wise multilayer perceptron (MLP). For each point $\mathbf x_j\in\mathbb R^3$, the MLP maps its coordinates to a $d_c$-dimensional feature $\mathbf f_j$. The MLP operates independently on each point, while anatomical and view identities are retained through the contour grouping.

Based on these identities, we extract the epicardial and endocardial features from SAX and LAX views as $\mathcal F^{r,m}\subseteq\mathcal F^{\mathrm{full}}$. Each feature set is independently pooled to obtain a slice-level representation $F^{r,m}$. The representations are further aggregated to obtain a global contour feature $G$.
\vspace{-2mm}

\subsubsection{Vertex-wise Local Feature Routing}

Reconstructing a 3D surface requires preserving the local relevance of individual contours while integrating information across observations. We introduce a Local Router that associates contour representations with each template vertex.

The template vertices $\mathcal V$ are projected to $d_c$-dimensional features $H=\{\mathbf h_i^0\}_{i=1}^{N_v}$. For each view $m$, a vertex-wise query $Q_i$ is projected from $\mathbf h_i^0$, and cross-attention with $F^{r,m}$ is computed as
\begin{equation}
Z_i^{r,m}
=
\operatorname{softmax}
\left(
\frac{Q_i(K^{r,m})^\top}{\sqrt{d_c}}
-
\lambda_d D_i^2
\right)
U^{r,m},
\label{eq:local_router}
\end{equation}
where $K^{r,m}$ and $U^{r,m}$ are the key and value projections of $F^{r,m}$. $D_i$ is the vertex-to-plane distance vector, with $D_i^2$ denoting its element-wise square, and $\lambda_d$ controls the distance bias. The resulting weights aggregate contour information for each vertex. The SAX- and LAX-derived contexts $Z_i^{r,\mathrm{SAX}}$ and $Z_i^{r,\mathrm{LAX}}$ are averaged and linearly projected to obtain $Z_i^r$.

\vspace{-4mm}
\subsubsection{Graph-based Template Deformation}

The Local Router provides vertex-specific contour information, which is propagated over the mesh through three Graph Blocks conditioned by Feature-wise Linear Modulation (FiLM)~\cite{perez2018film}.

The template features $H$ are progressively refined using the mesh connectivity $\mathcal E$. In the $\ell$-th Graph Block, each vertex feature is aggregated with the mean feature of its one-ring neighbors and layer-normalized, yielding $y_i^\ell$. The condition $c_i=G+Z_i^r$ modulates each block through FiLM, with a block-specific linear layer producing the scale and shift parameters:
\vspace{-1mm}
\begin{equation}
[\gamma_i^\ell;\beta_i^\ell]
=
\mathrm{Linear}^{\ell}(c_i).
\label{eq:film_parameters}
\end{equation}
The aggregated feature $y_i^\ell$ is then modulated by
\begin{equation}
\widetilde y_i^\ell
=
y_i^\ell\odot
\left[1+0.1\tanh(\gamma_i^\ell)\right]
+\beta_i^\ell,
\label{eq:film_modulation}
\end{equation}
where $\odot$ denotes element-wise multiplication. The modulated feature is transformed by GELU and a linear projection $O^\ell$, followed by a residual update:
\begin{equation}
h_i^{\ell+1}
=
h_i^\ell+
O^\ell\!\left(
\mathrm{GELU}(\widetilde y_i^\ell)
\right).
\label{eq:graph_update}
\end{equation}
After three Graph Blocks, an MLP predicts the 3D vertex displacements in Eq.~\eqref{eq:mesh_reconstruction}.

\subsection{Training Loss}
The training objective consists of a section loss $\mathcal L_{\mathrm{sec}}$, a volume loss $\mathcal L_{\mathrm{vol}}$, and a geometric regularization loss $\mathcal L_{\mathrm{reg}}$.

The section loss provides 2D supervision from the observed contours without requiring 3D mesh ground truth. For each acquisition plane $k$ and surface, the predicted mesh is intersected with the plane to obtain a cross-section $\hat{\mathcal S}_{k}^{r}$, which is compared with the corresponding contour $\mathcal C_{k}^{r}$ using bidirectional Chamfer distance in Eq.~\ref{eq:section_loss}, where $\Omega$ denotes the set of valid plane--surface pairs, $w_{k,r}$ is the normalized weight for each pair, and $D_{\mathrm{CD}}(\cdot,\cdot)$ denotes the bidirectional Chamfer distance.
\vspace{-2mm}
\begin{equation}
\mathcal L_{\mathrm{sec}}
=
\sum\nolimits_{(k,r)\in\Omega}
w_{k,r}
D_{\mathrm{CD}}\!\left(
\hat{\mathcal S}_{k}^{r},
\mathcal C_{k}^{r}
\right),
\label{eq:section_loss}
\end{equation}
For the volume loss, we apply a Huber penalty to the log volume ratio:
\begin{equation}
\mathcal L_{\mathrm{vol}}
=
H_{\eta}\!\left(
\log(\hat V / V)
\right),
\label{eq:volume_loss}
\end{equation}
where $\hat V$ and $V$ denote the reconstructed and reference volumes derived from the 2D annotations, respectively, and $H_{\eta}$ is the Huber loss with threshold $\eta$. 

We further employ a geometric regularizer $\mathcal L_{\mathrm{reg}}$ combining ARAP~\cite{sorkine2007arap}, edge-strain, and Laplacian regularization. The overall loss is
\vspace{-2.5mm}
\begin{equation}
\mathcal L
=
\mathcal L_{\mathrm{sec}}
+
0.25\mathcal L_{\mathrm{vol}}
+
\mathcal L_{\mathrm{reg}}.
\label{eq:training_objective}
\end{equation}

\vspace{-6.5mm}
\section{Experiments and Results}
\label{sec:Experiments}
\vspace{-2mm}

\subsection{Datasets}
We evaluate our method on two CMR datasets, M\&Ms-2~\cite{martinisla2023mnms2} and ACDC~\cite{bernard2018acdc}. M\&Ms-2 contains 360 subjects with both SAX and LAX acquisitions and is used for model development and reconstruction evaluation. ACDC contains 150 subjects with SAX acquisitions only and is used to evaluate cross-dataset generalization and downstream disease classification. Five-fold cross-validation is performed for all experiments. For both datasets, epicardial and endocardial contours are extracted from the manually annotated segmentation masks.

\vspace{-3mm}
\subsection{Evaluation Metrics}
We evaluate the reconstructed geometry in terms of geometric accuracy, clinical measurements, and downstream disease classification. \textbf{Geometric metrics:} Dice, mean contour distance (MCD)~\cite{bai2018automated}, and the 95th-percentile Hausdorff distance (HD), where MCD and HD are computed between mesh-plane intersections and reference contours. \textbf{Clinical metrics:} mean absolute errors (MAE) of end-diastolic volume (EDV), end-systolic volume (ESV), and left ventricular ejection fraction (LVEF). \textbf{Classification metrics:} accuracy (ACC), Macro-F1, and Macro-AUC, with macro metrics accounting for class imbalance.

\vspace{-2mm}
\subsection{Results on M\&Ms-2}
\label{sec:mnms_reconstruction}
\vspace{-1mm}
\subsubsection{Reconstruction Performance on M\&Ms-2}
\begin{figure}[h]
\centering
\includegraphics[width=\columnwidth]{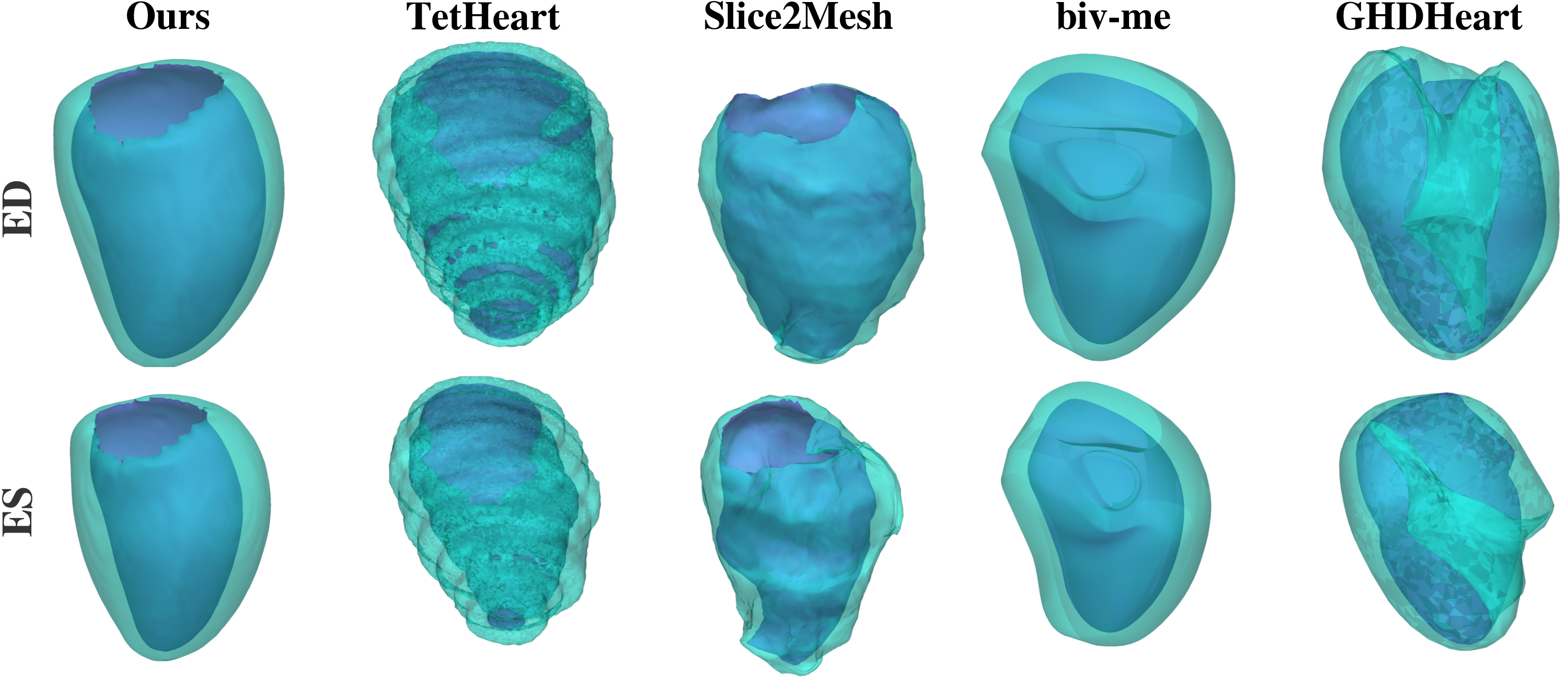}
\vspace{-8mm}
\caption{Qualitative comparison of LV reconstruction.}
\label{fig:mesh_comparison}
\end{figure}

\vspace{-1mm}
We first compare Local2Mesh with Slice2Mesh~\cite{xiao2025slice2mesh}, TetHeart~\cite{chen2025tetheart}, biv-me~\cite{dillon2025bivme}, and GHDHeart~\cite{luo2026ghdheart} on M\&Ms-2. As shown in Table~\ref{tab:mnms_reconstruction}, our method achieves the best geometric reconstruction, with the highest Mean Dice of $91.45\%$ and the lowest MCD and HD. For functional assessment, it achieves the lowest LVEF MAE. Fig.~\ref{fig:mesh_comparison} further shows that our method produces smoother and more coherent surfaces than competing methods.

\vspace{-3mm}
\begin{table}[h]
\centering    
\caption{LV reconstruction results on M\&Ms-2.}
\label{tab:mnms_reconstruction}
\footnotesize
\setlength{\tabcolsep}{2.2pt}
\renewcommand{\arraystretch}{0.92}

\begin{adjustbox}{max width=\columnwidth}
\begin{tabular}{lcccccc}
\toprule
\textbf{Method}
& \shortstack{\textbf{Mean Dice}\\\textbf{(\%) $\uparrow$}}
& \shortstack{\textbf{MCD}\\\textbf{(mm) $\downarrow$}}
& \shortstack{\textbf{HD}\\\textbf{(mm) $\downarrow$}}
& \shortstack{\textbf{EDV MAE}\\\textbf{(mL) $\downarrow$}}
& \shortstack{\textbf{ESV MAE}\\\textbf{(mL) $\downarrow$}}
& \shortstack{\textbf{LVEF MAE}\\\textbf{(\%) $\downarrow$}} \\
\midrule

Slice2Mesh
& 82.28
& 8.08
& 24.72
& 16.88
& 22.62
& 11.96 \\

TetHeart
& 87.09
& 2.43
& 6.37
& 15.9
& 10.00
& 6.07 \\

biv-me
& 89.69
& 5.47
& 19.19
& 14.76
& \textbf{3.63}
& 5.43 \\

GHDHeart
& 90.28
& 6.74
& 24.62
& 26.29
& 17.92
& 9.89 \\

Local2Mesh (Ours)
& \textbf{91.45}
& \textbf{1.70}
& \textbf{4.22}
& \textbf{7.90}
& 6.42
& \textbf{2.52} \\

\bottomrule
\end{tabular}
\end{adjustbox}
\end{table}

\vspace{-2mm}

\vspace{-5mm}
\subsubsection{Ablation Study on M\&Ms-2}
\vspace{-1mm}
We conduct ablation studies to evaluate the Local Router and FiLM-conditioned Graph Blocks. Removing the Local Router reduces Mean Dice by $2.22\%$ and increases MCD from $1.70$ to $2.06$ mm (Table~\ref{tab:ablation}). Replacing the Graph Blocks with graph convolutional network (GCN) layers~\cite{kipf2017semi} without FiLM conditioning (Plain GCN) reduces Mean Dice by $2.03\%$ and increases LVEF MAE from $2.52\%$ to $3.12\%$. Results demonstrate the contributions of local contour routing and conditioned graph deformation.

\vspace{-3mm}
\begin{table}[h]
\centering
\caption{Ablation study on M\&Ms-2.}
\label{tab:ablation}
\footnotesize
\setlength{\tabcolsep}{1.5pt}
\renewcommand{\arraystretch}{1.0}

\begin{adjustbox}{max width=\columnwidth}
\begin{tabular}{@{}lcccccc@{}}
\toprule
\textbf{Method}
& \shortstack{\textbf{Mean Dice}\\\textbf{(\%) $\uparrow$}}
& \shortstack{\textbf{MCD}\\\textbf{(mm) $\downarrow$}}
& \shortstack{\textbf{HD}\\\textbf{(mm) $\downarrow$}}
& \shortstack{\textbf{EDV MAE}\\\textbf{(mL) $\downarrow$}}
& \shortstack{\textbf{ESV MAE}\\\textbf{(mL) $\downarrow$}}
& \shortstack{\textbf{LVEF MAE}\\\textbf{(\%) $\downarrow$}} \\
\midrule

Ours
& \textbf{91.45}
& \textbf{1.70}
& \textbf{4.22}
& \textbf{7.90}
& \textbf{6.42}
& \textbf{2.52} \\

w/o Local Router
& 89.23
& 2.06
& 4.93
& 9.56
& 7.39
& 3.00 \\

Plain GCN
& 89.42
& 1.99
& 4.81
& 8.64
& 7.39
& 3.12 \\

\bottomrule
\end{tabular}
\end{adjustbox}
\end{table}

\vspace{-4mm}
\subsection{Evaluation on ACDC}
\label{sec:acdc}
\vspace{-1mm}
\subsubsection{Cross-Dataset Generalization on ACDC}

\vspace{-1mm}
We evaluate cross-dataset generalization on ACDC under zero-shot, training-from-scratch, and fine-tuning settings. Since ACDC provides only SAX acquisitions, we remove the LAX branch and directly use $Z^{r,\mathrm{SAX}}$ as the local representation $Z^r$. Since biv-me and GHDHeart optimize each case independently rather than learning a dataset-level model, zero-shot and fine-tuning are not applicable to these methods.

As shown in Table~\ref{tab:acdc_generalization}, our method achieves the best reconstruction performance across all applicable settings. In the zero-shot setting, it achieves a Mean Dice of $85.01\%$, exceeding TetHeart by $6.68$ percentage points, with lower MCD and HD. When trained from scratch, our method achieves the highest Mean Dice of $90.29\%$ and the lowest MCD, HD, ESV MAE, and LVEF MAE among all methods. Fine-tuning further improves Mean Dice to $91.17\%$ and reduces MCD and HD to $1.80$ and $3.89$ mm, respectively, outperforming Slice2Mesh and TetHeart across all metrics. 
\vspace{-3mm}
\begin{table}[h]
\centering
\caption{Reconstruction performance on ACDC.}
\label{tab:acdc_generalization}
\footnotesize
\setlength{\tabcolsep}{1.2pt}
\renewcommand{\arraystretch}{0.9}

\begin{adjustbox}{max width=\columnwidth}
\begin{tabular}{llcccccc}
\toprule
\textbf{Setting}
& \textbf{Method}
& \shortstack{\textbf{Mean Dice}\\\textbf{(\%) $\uparrow$}}
& \shortstack{\textbf{MCD}\\\textbf{(mm) $\downarrow$}}
& \shortstack{\textbf{HD}\\\textbf{(mm) $\downarrow$}}
& \shortstack{\textbf{EDV MAE}\\\textbf{(mL) $\downarrow$}}
& \shortstack{\textbf{ESV MAE}\\\textbf{(mL) $\downarrow$}}
& \shortstack{\textbf{LVEF MAE}\\\textbf{(\%) $\downarrow$}} \\
\midrule

\multirow{5}{*}{Zero-shot}
& Slice2Mesh & 56.50 & 10.24 & 24.60 & 58.15 & 92.43 & 59.75 \\
& TetHeart   & 78.33 & 4.57 & 11.20 & 23.78 & 25.52 & 15.72 \\
& biv-me     & -- & -- & -- & -- & -- & -- \\
& GHDHeart   & -- & -- & -- & -- & -- & -- \\
& Ours       & \textbf{85.01} & \textbf{3.00} & \textbf{6.11} & \textbf{10.31} & \textbf{18.90} & \textbf{8.12} \\

\midrule

\multirow{5}{*}{From scratch}
& Slice2Mesh & 72.16 & 3.95 & 13.07 & 86.40 & 51.16 & 39.79 \\
& TetHeart   & 87.20 & 2.43 & 5.73 & \textbf{11.10} & 13.80 & 7.01 \\
& biv-me     & 87.32 & 2.66 & 6.45 & 17.48 & 11.39 & 8.98 \\
& GHDHeart   & 88.85 & 3.78 & 16.01 & 42.09 & 24.24 & 9.42 \\
& Ours       & \textbf{90.29} & \textbf{1.96} & \textbf{4.06} & 13.35 & \textbf{9.83} & \textbf{3.97} \\

\midrule

\multirow{5}{*}{Fine-tuning}
& Slice2Mesh & 81.99 & 3.49 & 9.89 & 23.53 & 30.10 & 12.86 \\
& TetHeart   & 86.51 & 2.62 & 6.58 & 12.96 & 16.43 & 9.23 \\
& biv-me     & -- & -- & -- & -- & -- & -- \\
& GHDHeart   & -- & -- & -- & -- & -- & -- \\
& Ours       & \textbf{91.17} & \textbf{1.80} & \textbf{3.89} & \textbf{11.39} & \textbf{6.82} & \textbf{3.33} \\

\bottomrule
\end{tabular}
\end{adjustbox}
\end{table}
\vspace{-0.5mm}

\vspace{-5mm}
\subsubsection{Downstream Disease Classification}
\vspace{-1mm}
We further evaluate the utility of reconstructed 3D geometry for disease classification. Using the fine-tuned ACDC model, dense points are sampled from reconstructed meshes and compared with sparse points from aligned contours. Both are classified using PointNet~\cite{qi2017pointnet}, while CMR images with masks are classified using 3D ResNet-18~\cite{he2016resnet}. For all inputs, ED and ES features are encoded separately and concatenated for classification.

As shown in Table~\ref{tab:downstream_classification}, reconstructed meshes achieve $80.00\%$ ACC, $78.14\%$ Macro-F1, and $92.78\%$ Macro-AUC, outperforming both sparse contours and CMR images with masks. These results demonstrate the utility of reconstructed 3D geometry for downstream disease classification.
\vspace{-3mm}
\begin{table}[h]
\centering
\caption{Disease classification results on ACDC (\%).}
\label{tab:downstream_classification}
\small
\setlength{\tabcolsep}{1.8pt}
\renewcommand{\arraystretch}{0.95}
\begin{tabular*}{\columnwidth}{@{\extracolsep{\fill}}lcccc@{}}
\toprule
\textbf{Input / Classifier}
& \textbf{Params}
& \textbf{ACC $\uparrow$}
& \textbf{F1 $\uparrow$}
& \textbf{AUC $\uparrow$} \\
\midrule
Dense mesh + PointNet
& 4.00M
& \textbf{80.00}
& \textbf{78.14}
& \textbf{92.78} \\

Sparse contours + PointNet
& 4.00M
& 46.67
& 47.76
& 90.00 \\

MRI + masks + ResNet-18
& 33.43M
& 73.33
& 69.29
& 84.44 \\
\bottomrule
\end{tabular*}
\end{table}

\vspace{-4mm}
\section{Conclusion}

\vspace{-1mm}
In this work, we propose Local2Mesh for 3D LV reconstruction from sparse CMR slices without 3D mesh supervision. Local2Mesh addresses inter-slice misalignment through geometric contour alignment, associates contour features with template vertices using a plane-aware Local Router, and conditions graph-based deformation via FiLM. Experiments on two public datasets show improved reconstruction and functional estimation, while downstream classification supports the utility of reconstructed geometry. Future work will extend Local2Mesh to temporally consistent 4D reconstruction.

\section{Acknowledgment}

This work was supported by the National Natural Science Foundation of China (NSFC) under Grant No.82604003 and No.82603698. We acknowledge the M\&Ms-2 and ACDC challenge organizers for providing the datasets. 
\bibliographystyle{IEEEbib}
\bibliography{strings,refs}

\end{document}